\documentclass[conference]{IEEEtran}
\IEEEoverridecommandlockouts

\usepackage{cite}
\usepackage{amsmath,amssymb,amsfonts}
\usepackage{algorithmic}
\usepackage{graphicx}
\usepackage{textcomp}
\usepackage{booktabs}
\usepackage{xcolor}

\def\BibTeX{{\rm B\kern-.05em{\sc i\kern-.025em b}\kern-.08em
    T\kern-.1667em\lower.7ex\hbox{E}\kern-.125emX}}
    
\begin{document}

\title{Distinct Theta Synchrony across Speech Modes: Perceived, Spoken, Whispered, and Imagined\\
\thanks{This work was partly supported by Institute of Information \& Communications Technology Planning \& Evaluation (IITP) grant funded by the Korea government (MSIT) (No. RS-2019-II190079, Artificial Intelligence Graduate School Program (Korea University), No. RS-2021-II-212068, Artificial Intelligence Innovation Hub, and No. RS-2024-00336673, AI Technology for Interactive Communication of Language Impaired Individuals).}
}

\author{
\IEEEauthorblockN{Jung-Sun Lee}
\IEEEauthorblockA{\textit{Dept. of Artificial Intelligence} \\
\textit{Korea University} \\
Seoul, Republic of Korea \\
jungsun\_lee@korea.ac.kr}
\and
\IEEEauthorblockN{Ha-Na Jo}
\IEEEauthorblockA{\textit{Dept. of Artificial Intelligence} \\
\textit{Korea University} \\
Seoul, Republic of Korea \\ 
hn\_jo@korea.ac.kr}
\and
\IEEEauthorblockN{Eunyeong Ko}
\IEEEauthorblockA{\textit{Dept. of Artificial Intelligence} \\
\textit{Korea University}\\
Seoul, Republic of Korea \\
eunyeong\_ko@korea.ac.kr}
}

\maketitle

\begin{abstract}
Human speech production encompasses multiple modes such as perceived, overt, whispered, and imagined, each reflecting distinct neural mechanisms. Among these, theta-band synchrony has been closely associated with language processing, attentional control, and inner speech. However, previous studies have largely focused on a single mode, such as overt speech, and have rarely conducted an integrated comparison of theta synchrony across different speech modes.
In this study, we analyzed differences in theta-band neural synchrony across speech modes based on connectivity metrics, focusing on region-wise variations. The results revealed that overt and whispered speech exhibited broader and stronger frontotemporal synchrony, reflecting active motor–phonological coupling during overt articulation, whereas perceived speech showed dominant posterior and temporal synchrony patterns, consistent with auditory perception and comprehension processes. In contrast, imagined speech demonstrated a more spatially confined but internally coherent synchronization pattern, primarily involving frontal and supplementary motor regions. These findings indicate that the extent and spatial distribution of theta synchrony differ substantially across modes, with overt articulation engaging widespread cortical interactions, whispered speech showing intermediate engagement, and perception relying predominantly on temporoparietal networks.
Therefore, this study aims to elucidate the differences in theta-band neural synchrony across various speech modes, thereby uncovering both the shared and distinct neural dynamics underlying language perception and imagined speech.
\end{abstract}


\section{INTRODUCTION}

Brain-computer interface (BCI) serves as brain-driven communication pathways that convert neural signals into actionable inputs for external systems\cite{b1}. In recent years, active BCI has emerged as a next-generation control interface, offering speech-based interaction by directly harnessing the user’s cognitive states and intentions\cite{b2}. Various types of user input have been studied, including imagined speech\cite{b3}, visual imagery\cite{b4}, motor imagery\cite{b5,b6,b7}, and even handwriting recognition using neural representations, each presenting unique advantages and limitations.

Furthermore, mental state decoding paradigms have been explored to interpret cognitive and affective conditions from EEG, extending BCI applications beyond motor or linguistic domains toward more adaptive and context-aware interaction\cite{b8,b9}. Communication impairment in patients with severe paralysis or locked-in syndrome poses a fundamental challenge for maintaining interaction with the external world. BCIs have therefore emerged as an essential technology, enabling direct translation of neural activity into control commands without muscular involvement. Among various paradigms, speech-based BCIs offer an intuitive and human-like pathway for restoring communication by decoding the neural correlates of speech production and imagination.

Previous studies have primarily focused on individual speech modes, such as overt speech production, whispered articulation, or imagined speech, to identify their distinct neural signatures\cite{b10,b11,b12}. For instance, overt and whispered speech engage well-defined sensorimotor and premotor networks, whereas imagined speech primarily involves higher-order temporal and frontal regions associated with internal verbalization\cite{b13}. Although these findings have advanced our understanding of speech-related neural processes, they remain limited in scope to single-mode decoding, making it difficult to capture the continuum of speech control in naturalistic contexts

To develop more robust and realistic BCIs for speech restoration, it is crucial to integrate and compare multiple speech modes within a unified experimental framework\cite{b14, b15, b16}. Such an approach enables the identification of both shared and mode-specific neural dynamics, providing insights into how the brain transitions from perception to articulation and imagination\cite{b17}. Despite the growing interest in this direction, few studies have systematically examined neural synchrony across all major speech modes using connectivity-based analyses\cite{b19, b18}.

\begin{figure*}[htp]
\centering
\scriptsize
\includegraphics[width=\textwidth]{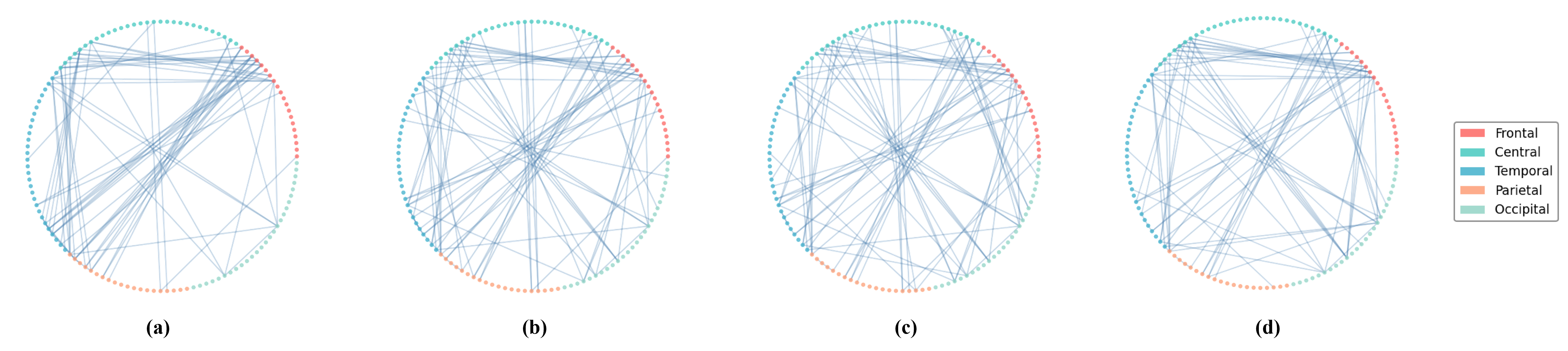}
\caption{Theta-band PLV-based functional connectivity across four speech modes: 
(a) perceived, (b) overt, (c) whispered, and (d) imagined speech. 
Each circular graph represents inter-regional phase synchrony derived from the theta band. 
Node colors indicate cortical regions (frontal, central, temporal, parietal, and occipital), 
as illustrated in the accompanying legend.}
\label{fig:theta_plv}
\end{figure*}

In this study, we investigated the differences in theta band neural synchrony across four speech modes: perceived, overt, whispered, and imagined, using connectivity metrics derived from high density electroencephalogram (EEG). We focused on cortical variations to reveal differential interactions between temporal and frontal regions, which play key roles in language perception and production. By elucidating the shared and distinct neural characteristics underlying these modes, our findings contribute to a deeper understanding of speech-related brain dynamics and provide foundational insights for the development of adaptive, speech-mode–aware BCI systems. To further strengthen the ecological validity of our findings, we also conducted exploratory analyses of inter-individual variability across speech modes, assessing how individual differences in cognitive state, attention, and neural baseline activity modulate theta-band synchrony patterns. Moreover, the features extracted from this frequency band can serve as effective representations for feature engineering in subsequent machine learning and deep learning analyses\cite{b20, b21}.

\section{MATERIALS AND METHODS}

\subsection{Dataset Description}
The study involved ten healthy participants, six males and four females, with a mean age of 23.7 years (SD = 3.23). None of the participants had a history of claustrophobia or hearing impairment, ensuring that the experimental conditions did not induce discomfort or auditory bias. Brain signals were acquired using a 128-channel EEG cap with active Ag/AgCl electrodes that followed the international 10-5 system for electrode placement. The FCz and FPz channels were designated as the reference and ground electrodes. EEG signals were captured using Brain Vision/Recorder software (Brain Products GmbH, Germany) and processed using MATLAB 2023a. A sampling rate of 1,000 Hz was employed to capture EEG recordings, facilitating the detailed examination of dynamic processes inherent in different speech paradigms. This study was approved by the Institutional Review Board at Korea University (KUIRB-2022-0104-04), and all participants provided written informed consent prior to participation.

\subsection{Signal Preprocessing}
EEG signals were analyzed using MNE-Python and MATLAB with BBCI toolbox. The EEG data were subjected to a series of pre-processing steps to ensure high-quality signals for analysis. First, a band-pass filter was applied to retain frequencies between 0.5 and 125 Hz\cite{b22}. A notch filter was also implemented at 60 Hz and 120 Hz to remove power line noise and its harmonics. Channels with poor signal quality, such as those affected by electrooculography artifacts, were identified and rejected. The data were then segmented into 1.5-second epochs without overlap, resulting in a total of 1,200 trials. These trials were evenly distributed across 20 classes, with 60 trials per class. Following the short-time Fourier transform (STFT) process, we extracted band power features by summing the power within specific frequency bands: the delta (\(\delta\), 1--4 Hz), theta (\(\theta\), 4--8 Hz), alpha (\(\alpha\), 8--12 Hz), beta (\(\beta\), 12--30 Hz), and gamma (\(\gamma\), 30--45 Hz) bands. This procedure generated a feature matrix of dimensions (number of time windows $\times$ 5 bands) for each EEG channel. 

\subsection{Functional Connectivity}

Extracting features that accurately capture the distinctive connectivity characteristics of each paradigm is crucial for achieving task-relevant representation learning\cite{b23, b24}. To quantify the functional connectivity across various frequency bands during perceived, overt, whispered, and imagined speech paradigms, we employed three metrics: phase locking value (PLV)\cite{b25} and phase lag index (PLI)\cite{b26}, and coherence. By constructing networks using these three metrics, we investigated the different connection patterns among the four speech patterns.

PLV was utilized to measure the average phase difference between pairs of EEG time series. The instantaneous phases of the signals were obtained using the Hilbert transform. For signals \( x_n \) and \( x_t \) at time point \( k \), the PLV is defined as:

\begin{equation}
PLV_{n,t} = 
\left|\frac{1}{M}\sum_{k=0}^{M-1}
e^{i(\phi_n(k)-\phi_t(k))}\right|,
\tag{1}\label{eq:plv}
\end{equation}

\begin{figure*}[htp]
\centering
\scriptsize
\includegraphics[width=\textwidth]{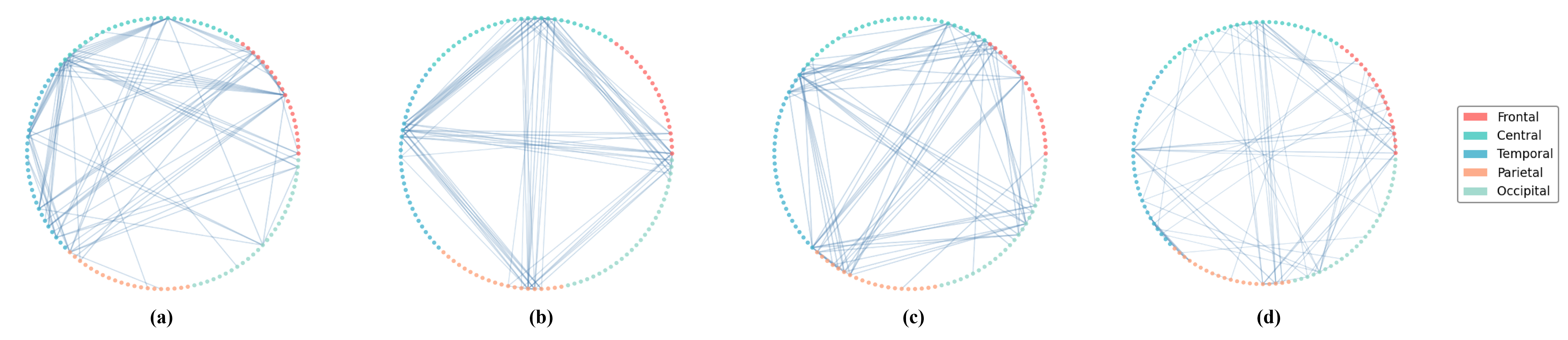}
\caption{Theta-band coherence-based functional connectivity across four speech modes: 
(a) perceived, (b) overt, (c) whispered, and (d) imagined speech. 
Each circular graph represents inter-regional phase synchrony derived from the theta band. 
Node colors indicate cortical regions (frontal, Ceftral, temporal, parietal, and occipital), 
as illustrated in the accompanying legend.}
\label{fig:theta_coh}
\end{figure*}

\noindent where \( \phi_n(k) \) and \( \phi_t(k) \) represent the instantaneous phases of signals \( x_n \) and \( x_t \) at time point \( k \), respectively, and \( M \) is the total number of samples. The PLV ranges from 0 to 1, with values closer to 1 indicating strong phase synchronization between the signals. This measure provides insight into the temporal coherence and functional connectivity between different EEG channels.

PLI was employed to quantify the average phase lead or lag between two EEG time series by analyzing the sign function of phase differences. Specifically, the PLI between signals \( x_n \) and \( x_t \) at time point \( k \) is defined as:

\begin{equation}
PLI_{n,t} = 
\frac{1}{M}\sum_{k=0}^{M-1}
\operatorname{sgn}\!\left(\phi_n(k)-\phi_t(k)\right),
\tag{2}\label{eq:pli}
\end{equation}

\noindent where the \( \text{sgn} \) function assigns a value of -1, 0, or 1 based on whether the phase difference is negative, zero, or positive, respectively. The PLI ranges from 0 to 1, with values closer to 1 indicating significant phase lead or lag synchronization between the signals. Unlike PLV, PLI is insensitive to common sources and volume conduction, making it a robust measure for assessing true functional connectivity.

The magnitude-squared coherence between two EEG signals \( x_n \) and \( x_t \) is defined as:

\begin{equation}
\mathrm{Coh}_{n,t}(f) =
\frac{|\mathcal{S}_{n,t}(f)|^2}
{\mathcal{S}_{n,n}(f)\cdot\mathcal{S}_{t,t}(f)},
\tag{3}\label{eq:coh_w}
\end{equation}

\noindent where \( \mathcal{S}_{n,t}(f) \) denotes the cross-spectral density between \( x_n \) and \( x_t \) at frequency \( f \), and \( \mathcal{S}_{n,n}(f) \) and \( \mathcal{S}_{t,t}(f) \) represent their respective auto-spectral densities. The coherence value ranges from 0 to 1, with values near 1 indicating a high degree of linear coupling between the two signals at frequency \( f \).

Coherence was computed for each pair of EEG channels within each frequency band to form frequency-specific connectivity matrices. This measure captures the strength and stability of oscillatory coupling, offering insight into how neural populations interact and synchronize during perceived, overt, whispered, and imagined speech paradigms. 

\section{RESULTS AND DISCUSSION}
Fig. 1 presents the group-averaged theta-band PLV connectivity maps for each speech mode, revealing distinct large-scale synchrony patterns across perception and articulation states. During passive perception of auditory stimuli, connectivity was strongest in bilateral temporal and posterior cortical regions, reflecting the dominance of auditory–perceptual processing. Positive mean differences, where imagined speech connectivity exceeded perceived speech, were primarily observed between frontal and prefrontal electrodes, including pairs such as AF8–F10, Fp1–F10, F4–Fp2, and AF7–F10. Increased connectivity was also identified in temporal-frontal pairs such as T8–FFT8h. 
Fig. 2 presents the group-averaged theta-band coherence connectivity maps for each speech mode, revealing distinct large-scale synchrony patterns across perception and articulation states.
Theta synchrony was particularly evident between temporal and parietal electrodes, suggesting stable engagement of the auditory comprehension network with limited anterior involvement.

In overt articulation, connectivity extended broadly across the frontal, central, and temporal areas, indicating strong coupling between sensorimotor and auditory cortices.
Enhanced fronto-temporal and central synchrony implies the integration of motor planning, vocal execution, and auditory feedback processes required for speech production. Whispered articulation exhibited an intermediate topology, showing preserved but weakened fronto-central connections compared to overt speech.
The reduced global coherence and partial maintenance of anterior synchrony suggest that motor coordination remains active without full vocal output, representing a transitional state between overt and imagined articulation.

In the imagined condition, overall theta connectivity decreased markedly, particularly in prefrontal and motor-related regions.
Residual coupling persisted in posterior and temporal networks, implying reliance on internally simulated perceptual representations rather than overt motor execution. Table \ref{tab:regionwise_theta_overt_imagined} summarizes the region-wise connectivity counts and mean weights across frequency bands, showing that the strongest connectivity differences between overt and imagined speech occurred in the theta band, particularly within and between frontal and central regions. Therefore, leveraging these characteristic patterns could enable the development of integrative systems that unify multiple BCI interfaces across diverse environments\cite{b27, b28}.

\begin{table}[t]
    \centering
    \caption{Region-wise connectivity counts for overt and imagined speech conditions}
    \label{tab:regionwise_theta_overt_imagined}
    \renewcommand{\arraystretch}{1.2}
    \begin{tabular}{lccc}
        \toprule
        \textbf{Region Pair} & \textbf{Connections} & \textbf{Mean Weight} & \textbf{Frequency} \\
        \midrule
        Frontal $\leftrightarrow$ Parietal    & 166 & $-0.062$  & $\delta$ \\
        Parietal $\leftrightarrow$ Parietal   & 166 & $-0.064$  & $\delta$ \\
        Central $\leftrightarrow$ Parietal    & 163 & $-0.062$  & $\delta$ \\
        Frontal $\leftrightarrow$ Occipital   & 59  & $-0.069$  & $\delta$ \\
        Central $\leftrightarrow$ Central     & 146 & $-0.057$  & $\delta$ \\
        \midrule
        Frontal $\leftrightarrow$ Frontal      & 972 & $-0.103$ & $\theta$ \\
        Frontal $\leftrightarrow$ Parietal     & 344 & $-0.059$ & $\theta$ \\
        Frontal $\leftrightarrow$ Central      & 311 & $-0.073$ & $\theta$ \\
        Central $\leftrightarrow$ Parietal     & 298 & $-0.044$ & $\theta$ \\
        Central $\leftrightarrow$ Central      & 290 & $-0.060$ & $\theta$ \\
        \midrule
        Frontal $\leftrightarrow$ Central    & 141 & $0.076$ & $\alpha$ \\
        Frontal $\leftrightarrow$ Parietal   & 143 & $0.074$ & $\alpha$ \\
        Parietal $\leftrightarrow$ Parietal  & 90  & $0.063$ & $\alpha$ \\
        Central $\leftrightarrow$ Parietal   & 73  & $0.072$ & $\alpha$ \\
        Occipital $\leftrightarrow$ Central  & 41  & $0.087$ & $\alpha$ \\
        \midrule
        Frontal $\leftrightarrow$ Frontal    & 84  & $-0.034$ & $\beta$ \\
        Frontal $\leftrightarrow$ Prefrontal & 14  & $-0.042$ & $\beta$ \\
        Parietal $\leftrightarrow$ Parietal  & 4   & $0.020$  & $\beta$ \\
        Parietal $\leftrightarrow$ Central   & 2   & $0.024$  & $\beta$ \\
        Frontal $\leftrightarrow$ Temporal   & 1   & $-0.028$ & $\beta$ \\
        \midrule
        Frontal $\leftrightarrow$ Frontal      & 238 & $-0.047$ & $\gamma$ \\
        Prefrontal $\leftrightarrow$ Frontal   & 40  & $-0.068$ & $\gamma$ \\
        Frontal $\leftrightarrow$ Parietal     & 54  & $-0.023$ & $\gamma$ \\
        Frontal $\leftrightarrow$ Temporal     & 23  & $-0.031$ & $\gamma$ \\
        Parietal $\leftrightarrow$ Frontal     & 54  & $-0.023$ & $\gamma$ \\
        \bottomrule
    \end{tabular}
\end{table}

\section{CONCLUSIONS}
This study investigated theta-band neural dynamics underlying different speech paradigms using EEG-based decoding and statistical connectivity analysis. The results revealed that theta-band activity exhibited distinct synchronization patterns across speech states, particularly during imagined speech, where internal speech generation elicited unique cortical interactions not observed in overt or whispered conditions. Statistical comparisons confirmed significant theta-band differences, indicating that imagined speech recruits specialized oscillatory processes that diverge from externally expressed speech modes. Connectivity metrics such as PLV and PLI provided robust indicators of functional coupling within theta networks, capturing coherent neural activity across distributed cortical regions. These findings suggest that theta-band synchrony serves as a critical substrate for speech-mode differentiation and may reflect hierarchical coordination between perceptual and motor systems. Moreover, the analysis demonstrates that theta-band coupling strength can predict inter-mode variability, offering a neurophysiological basis for adaptive modeling. 

\bibliography{Reference}
\bibliographystyle{IEEEtran}

\end{document}